\documentclass[letterpaper, 10 pt, conference]{ieeeconf}  

\IEEEoverridecommandlockouts                              

\usepackage{graphics} 
\usepackage{graphicx} 
\usepackage{amsmath} 
\usepackage[noend]{algpseudocode}
\usepackage{algorithmicx,algorithm}
\usepackage{cite}
\usepackage{multirow}
\usepackage{subfigure}
\usepackage{hyperref}

\title{\LARGE \bf
Visual Manipulation Relationship Network
}

\author{Hanbo Zhang, Xuguang Lan, Xinwen Zhou, Zhiqiang Tian, Yang Zhang and Nanning Zheng
\thanks{*This work was supported in part by NSFC No. 91748208, the National Key Research and Development Program of China under grant No. 2017YFB1302200 and 2016YFB1000903,  NSFC No. 61573268, and Shaanxi Key Laboratory of Intelligent Robots.}
\thanks{Hanbo Zhang and Xuguang Lan are with the Institute of Artificial Intelligence, the National Engineering Laboratory for Visual Information Processing and Applications, School of Electronic and Information Engineering,
        Xi'an Jiaotong University, No.28 Xianning Road, Xi'an, Shaanxi, China.
        {\tt\small zhanghanbo163@stu.xjtu.edu.cn, xglan@mail.xjtu.edu.cn}}%
}

\begin{document}

\maketitle
\thispagestyle{empty}
\pagestyle{empty}

\begin{abstract}

Robotic grasping is one of the most important fields in robotics and convolutional neural network (CNN) has made great progress in detecting robotic grasps. However, including multiple objects in one scene can invalidate the existing grasping detection algorithms based on CNN because of lacking of manipulation relationships among objects to guide the robot to grasp things in the right order. Therefore, the manipulation relationships are needed to help robot better grasp and manipulate objects. This paper presents a new CNN architecture called Visual Manipulation Relationship Network (VMRN) to help robot detect targets and predict the manipulation relationships in real time. To implement end-to-end training and meet real-time requirements in robot tasks, we propose the Object Pairing Pooling Layer (OP$^2$L) to help to predict all manipulation relationships in one forward process. In order to train VMRN, we collect a dataset named Visual Manipulation Relationship Dataset (VMRD) consisting of 5185 images with more than 17000 object instances and the manipulation relationships between all possible pairs of objects in every image, which is labeled by the manipulation relationship tree. The experiment results show that the new network architecture can detect objects and predict manipulation relationships simultaneously and meet the real-time requirements in robot tasks.
 
 \end{abstract}

\section{Introduction}

\begin{figure}[tb] 
 \center{\includegraphics[scale=0.32]{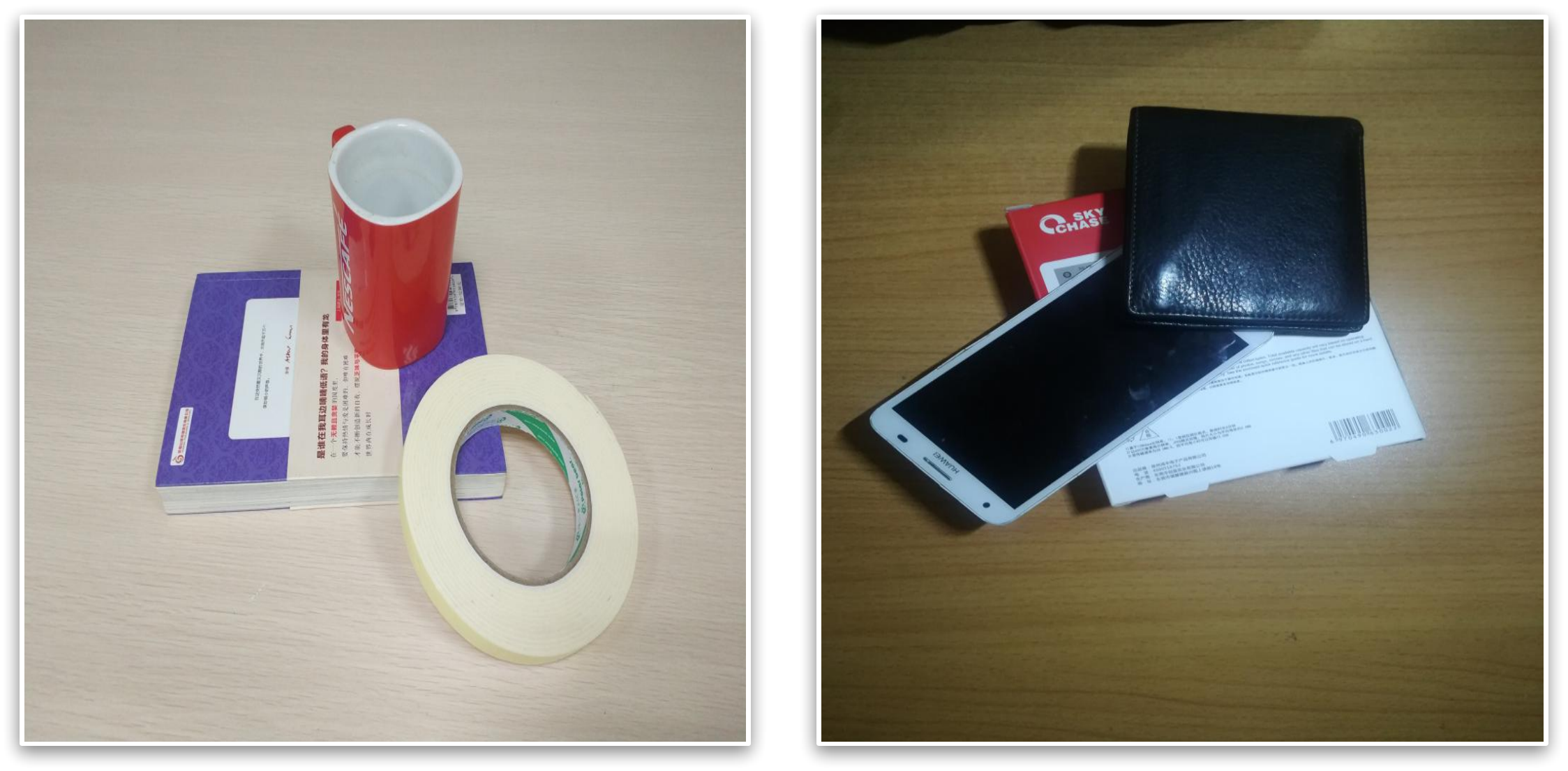}}        
 \caption{Importance of manipulation order. Left: A cup is on a book. Right: A phone is on a box. As shown in two scenes, if we do not consider the relationships of manipulation, and robots choose to pick up the book or the box first, then the cup or the phone may be dumped or even broken. } 
 \label{fig1}
 \end{figure}

Grasping is one of the most significant manipulation in everyday life. Robotic grasping develops rapidly in recent years. However, it is still far behind human performance and remains unsolved. For example, when humans encounter a stack of objects like shown in Fig. \ref{fig1}, they instinctively know how to grasp them. As for the robot, when it is planning to grasp a stack of objects, the process includes detecting objects and their grasps, determining the grasping order and finally planning grasping motion. These problems still remain challenging and hinder the widespread use of robots in everyday life.
 
Recently, with rapid development of deep learning, it is proved to be a useful tool in computer vision, which has made impressive breakthroughs in many research fields, such as image classification\cite{imagenet} and object detection\cite{SSD,c5,c4}. The reason that deep learning has made so much progress is that through its training, deep networks can extract features of objects or images that are far superior to hand-designed ones \cite{DL}. These advantages have encouraged researchers to apply deep learning to robotics. Some recent works have proved the effectiveness of deep learning and convolutional neural network (CNN) in robotic perception \cite{spatialae,handeyegrasp} and control \cite{opendoor,endtoend}. In particular, deep learning has achieved unprecedented performance in robotic grasp detection \cite{lenz,redmongrasp,selfsupervisedgrasp,resnetgrasp}. Most current robotic grasp detection methods take RGB or RGB-D images as input and output a vectorized and standardized grasps. These robotic grasp detection algorithms essentially solve an object detection problem. They treat the grasps as a special kind of objects and detect them by using a neural network. 

However, using this type of grasp detection algorithm for robotic grasping experiments can only deal with scenes containing a single target. Robot will execute the grasp having the highest confidence score. Doing this can have a devastating effect on objects in some multi-object scenes. For example, as shown in Fig. \ref{fig1}, a cup is placed on a book, and if the detected grasp with the highest confidence score belongs to the book, which means the robot chooses to pick up the book first, the cup may fall apart and be broken. Therefore, in this paper, we focus on helping the robot find the right grasping order when it is facing a stack of objects, which is defined as {\bf manipulation relationship prediction}.

Some recent works have used convolutional neural network to predict the relationships between objects rather than just object detection\cite{c1,c2,rel1,rel2}. These works show that convolutional neural network has the potential to understand the relationships between objects. Therefore, we hope to establish a method based on neural network so that the robot can understand the manipulation relationships between objects in multi-object scenes to help the robot finish more complicated grasp tasks.

In our work, we design a new network architecture named Visual Manipulation Relationship Network (VMRN) to simultaneously detect objects and predict the manipulation relationships. The network architecture has two stages. The output of the first-stage is the object detection result, and the output of the second-stage is the prediction result of the manipulation relationships. To train our network, we contribute a new dataset called Visual Manipulation Relationship Dataset (VMRD). The dataset contains 5185 images of hundreds of objects with 51530 manipulation relationships and the category and location information of each object. The format of the annotations refers to the PASCAL VOC dataset. In summary, the contributions of our work include three points:

\begin{itemize}
\item We design a new convolutional neural network architecture to simultaneously detect objects and predict manipulation relationships, which meets the real-time requirements in robot tasks.
\item We collect a new dataset of hundreds graspable objects, which includes the location and category information and the manipulation relationships between pairs of objects.
\item As we know, it is the first end-to-end architecture to predict robotic manipulation relationships directly using images as input with convolutional neural network.
\end{itemize}

The rest part of this paper is organized as follows: section II reviews the background and related works; section III introduces the problem formulation including the object detection and representation of manipulation relationships; section IV gives the details of our approach and network including the training method; section V shows the experiment results of manipulation relationship prediction and image-based and object-based test including some subjective experiment results; and finally, the conclusion of our results and future work are discussed in section VI.

\section{Background}

\subsection{Object Detection}

Object detection is defined as a process using a image including several objects as input to locate and classify as many target objects as possible in the image. Sliding window used to be the most common method to detect objects. When using such way to do object detection, the features, such as HOG\cite{HOG} and SIFT\cite{SIFT}, of the target objects are usually extracted first, and then they are used to train a classifier, like Supported Vector Machine, to classify the candidates coming from sliding window stage. Deformable Parts Model (DPM)\cite{DPM} is the most successful one of this type of object detection algorithms. 

Recently, object detection algorithms based on deep features, such as Region-based CNN (RCNN) family\cite{c4,c5} and Single Shot Detector (SSD) family\cite{SSD}, are proved to drastically outperform the previous algorithms which are based on hand-designed features. Based on the detection process, the main algorithms are classified into two types, which we call one-stage algorithms such as SSD\cite{SSD} and two-stage algorithms such as Faster RCNN\cite{c5}. One-stage algorithms are usually faster than two-stage algorithms while two-stage algorithms often get better results \cite{objdetecreview}.

Our work focuses on not only the object detection, but also the manipulation relation prediction. The challenge is how to combine the relation prediction stage with object detection stage. To solve this problem, we design the Object Pairing Pooling Layer, which is used to generate the input of manipulation relationship predictor using the object detection results and convolutional feature maps as input. The details will be described in following sections.
 
 \subsection{Visual Relationship Detection}
 
Visual relationship detection means understanding object relationships of an image. Some previous works try to learn spatial relationships\cite{relseg,relspatial}. Later, researchers attempt to collect relationships from images and videos and help models map these relationships from images to language\cite{rel3,rel4,rel5,rel6}. Recently, with the help of deep learning, the relationship prediction between objects has made a great process\cite{c1,c2,rel1,rel2}. Lu et al. \cite {c1} collect a new dataset of object relationships called Visual Relationship Dataset and propose a new relationship prediction model consisting of visual and language parts, which outperforms previous methods. Liang et al. \cite {c2} firstly combine deep reinforcement learning with relationships and their model can sequentially output the relationships between objects in one image. Yu et al. \cite{rel1} use internal and external linguistic knowledge to compute the conditional probability distribution of a predicate given a $(subject,object)$ pair, which achieves better performance. Dai et al. \cite{rel2} propose an integrated framework called Deep Relational Network for exploiting the statistical dependencies between objects and their relationships.

These works focus on relationships represented by linguistic information between objects but not manipulation relationships. In our work, we introduce relationship detection methods to help robot find the right order in which the objects should be manipulated. And because of the real-time requirements of robot system, we need to find a way to accelerate the prediction of manipulation relationships. Therefore, we propose an end-to-end architecture different from all previous works.

\section{Problem Formulation}

   \begin{figure}[b]        
 \center{\includegraphics[scale=0.25]{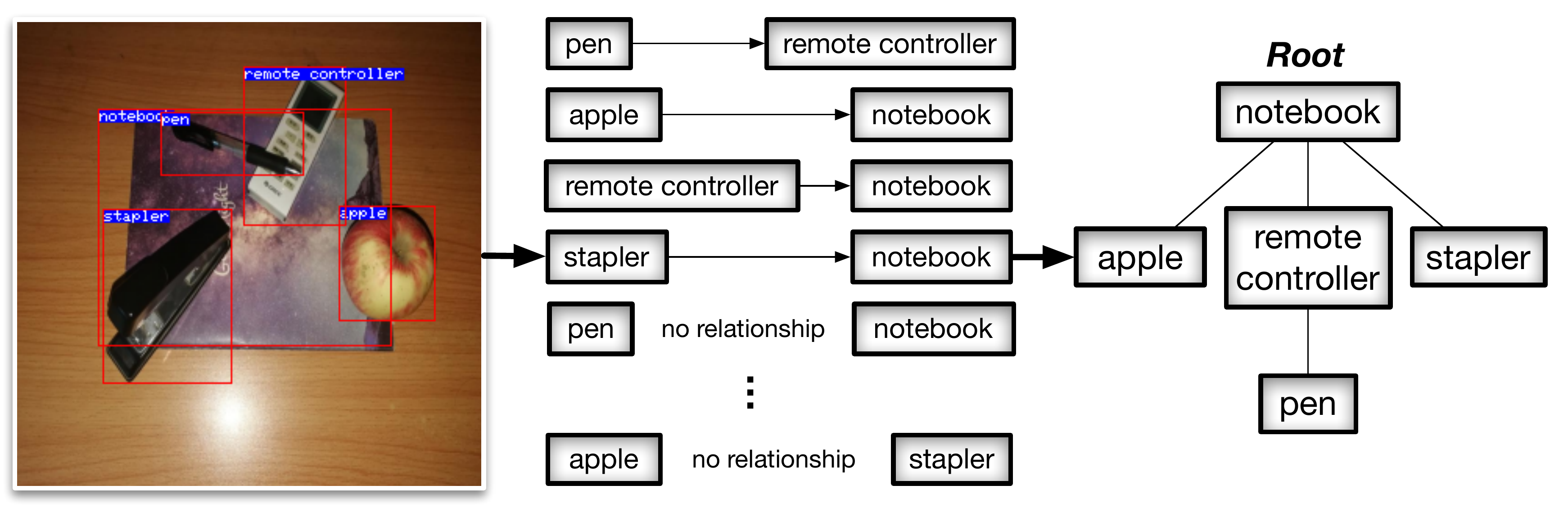}}        
 \caption{An example of manipulation relationship tree. Left: Images including several objects. Middle: All pair of objects and manipulation relationships. Right: manipulation relationship tree, in which the leaf nodes should be manipulated before the other nodes.}   
 \label{fig2}   
 \end{figure}

   \begin{figure*}[tb]  
 \center{\includegraphics[scale=0.32]{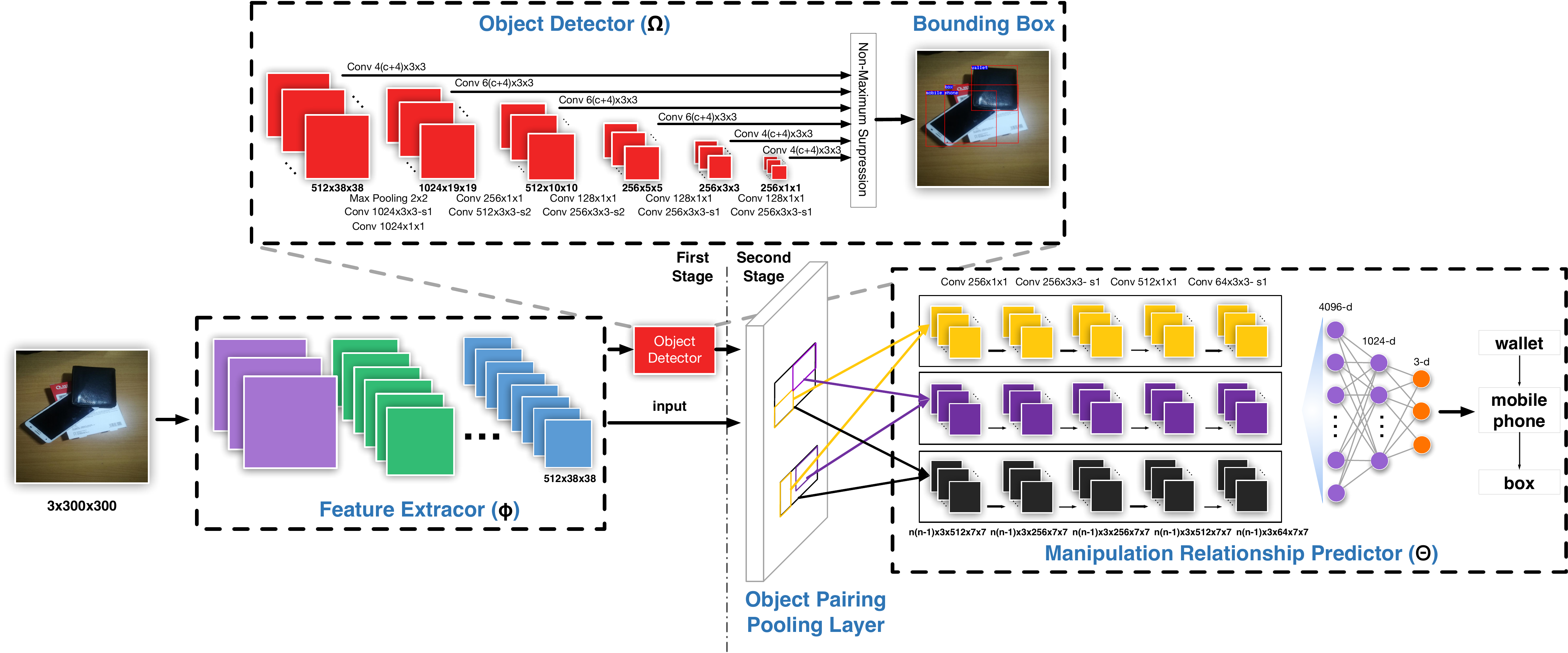}}        
 \caption{Network architecture of VMRN. Input of the network is $3 \times 300 \times 300$ images including several graspable objects. Feature extractor is a stack of convolution layers ($e.g.$ VGG \cite{c6} or ResNet \cite{c7}), which output feature maps with size of $512 \times 38 \times 38$. These features are used by object detector and OP$^2$L to respectively detect objects and generate the feature groups of all possible object pairs which are used to predict manipulation relationships by manipulation relationship predictor.}  
 \label{fig3} 
 \end{figure*}
 
\subsection{Object Detection} 
 
Output of object detection is the location of each object and its category. The location of objects is represented by a vertical bounding box, which is a 4-dimensional vectors: ${loc} = (x_{min}, y_{min}, x_{max}, y_{max})$, where $(x_{min}, y_{min})$ and $(x_{max}, y_{max})$ represent the coordinates of upper left vertex and lower right vertex, respectively. The category is encoded as an integer which is the index of the maximum value of classification confidence scores:  ${cls} = index(max(conf_{1},...,conf_{c}))$, where $c$ is the number of categories and $(conf_{1},...,conf_{c})$ is the confidence score vector with each element standing for the likelihood that the object is classified into the corresponding category.
 
\subsection{Manipulation Relationship Representation}
 
In this paper, manipulation relationship is the order of grasping. Therefore, we need a objective criterion to determine the grasping order, which is described as following: if moving one object will have an effect on the stability of another object, this object should not be moved first. Since we only focus on the manipulation relationships between objects and do not concern the linguistic information, a tree-like structure (two objects may have one same child), called \textbf {manipulation relationship tree} in the following, can be constructed to represent the manipulation relationships of all the objects in each image. Objects are represented by nodes and parent-child relationships between nodes indicate the manipulation relationships. In manipulation relationship tree, the object represented by the parent node should be grasped after the object represented by the child node. Fig. \ref{fig2} is an example of the manipulation relationship tree. A pen is on a remote controller, and a remote controller, an apple and a stapler are on a book. Therefore, the pen is the child of the remote controller and the remote controller, the apple and the stapler are children of the book in the manipulation tree.

\section{Proposed Approach}
 
The proposed network architecture is shown in Fig. \ref{fig3}. The inputs of our network are images and outputs are object detection results and manipulation relationship trees. Our network consists of three parts: feature extractor, object detector and manipulation relationship predictor, with parameters denoted by $\Phi$, $\Omega$ and $\Theta$ respectively.

In our work, taking into account the real-time requirements of object detection, we use Single Shot Detector (SSD) algorithm\cite{SSD} as our object detector. SSD is an one-stage object detection algorithm based on CNN. It utilizes multi-scale feature maps to regress and classify bounding boxes in order to adapt to object instances with different size. Input of object detector is convolution feature maps (in our work, we use VGG \cite {c6} or ResNet50 \cite {c7} features). Through object classification and multi-scale object location regression, we obtain the final object detection results. The result of each object is a 5-dimensional vector $(cls, x_{min}, y_{min}, x_{max}, y_{max})$. Then the inputs of Object Pairing Pooling Layer (OP$^2$L) are object detection results and convolution features. The outputs are concatenated as a mini-batch for predicting manipulation relationships by traversing all possible pairs of objects. Finally, the manipulation relationship between each pair of objects is predicted by manipulation relationship predictor.
 
\subsection{Object Pairing Pooling Layer}

OP$^2$L is designed to implement the end-to-end training of the whole network. In our work, weights of feature extractor $\Phi$ are shared by manipulation relationship predictor and object detector. OP$^2$L is added between feature extractor and manipulation relationship predictor like in Fig. \ref{fig3}, using object location ($e.g.$ the online output of object detection or the offline ground truth bounding box) and shared feature maps $CNN(I;\Phi)$ as input, where $I$ is the input image. It finds out all possible pairs ($n$ objects correspond to $n(n-1)$ pairs) of objects and makes their features a mini-batch to train the manipulation relationship predictor. Although in complex visual relationship prediction tasks, traversing all possible object pairs is time-consuming \cite {relpn} due to the large number of objects in the scene and the sparsity of the relationships between the objects. However, in our manipulation relationship prediction task, there are only a few of objects in the scene and it does not take a long time to traverse all the object pairs.
 
Let $O_i$ and $O_j$ stand for an object pair.  OP$^2$L can generate the features of $O_i$ and $O_j$ denoted by $CNN(O_i,O_j;\Phi)$, which includes features of two objects and their union. In detail, the features are cropped from shared feature maps and adaptively pooled into a fixed spatial size $H \times W$ ($e.g. 7 \times 7$). The gradients with respect to the same object or union bounding box coming from manipulation relationship predictor are accumulated and propagated backward to the front layers.

\begin{figure}[tb]        
 \center{\includegraphics[scale=0.35]{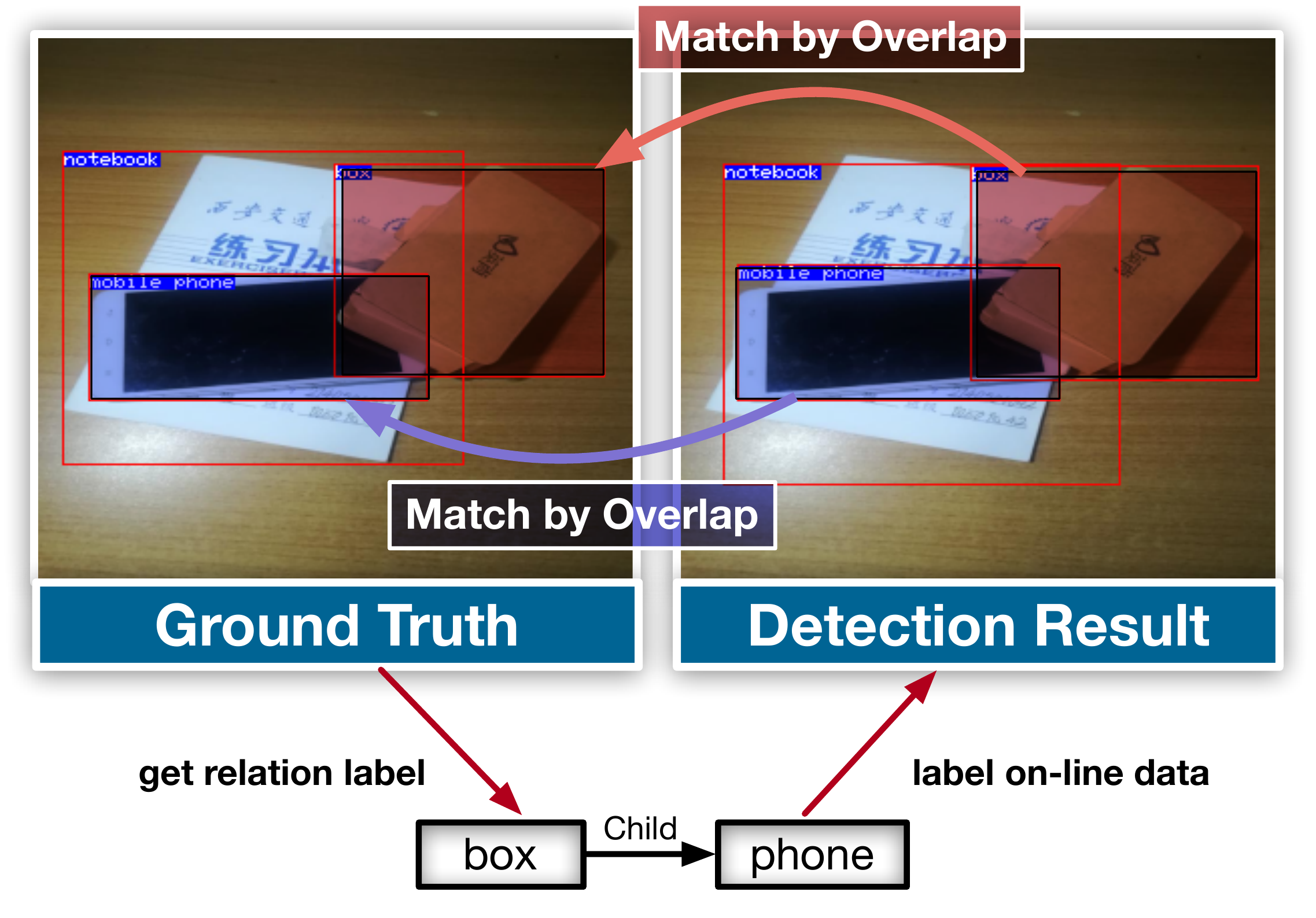}}        
 \caption{Method to label online data. First, we match the predicted bounding boxes to the ground truth by areas of overlap. Then we use the manipulation relationship between ground truth bounding boxes as the ground truth manipulation relationship between predicted bounding boxes to generate online data used to train manipulation relationship predictor. } 
 \label{fig4}  
 \end{figure}
 
\subsection{Training Data of Relation Predictor} 

An extra branch of CNN is cascaded after OP$^2$L to predict manipulation relationships between objects. Training data for manipulation relationship predictor $D_{RP}$ is generated by OP$^2$L, which includes two parts: online data $D_{on}$ and offline data $D_{off}$, coming from object detection results and ground truth bounding boxes respectively. That is to say $D_{RP}=D_{on} \cup D_{off}$. For each image, $D_{RP}$ is a set of CNN features $CNN(O_i,O_j;\Phi)$ of all possible object pairs and their labels $(O_i,R,O_j)$, where $R$ is the manipulation relationship between $O_i$ and $O_j$. The reason we mix online data and offline data to train manipulation relationship predictor is that online data can be seen as the augmentation of offline data while offline data can be seen as the correction of online data. Manipulation relationships between online object instances are labeled according to the manipulation relationships between ground truth bounding boxes that maximumly overlap the online ones. As shown in Fig. \ref{fig4}, object detection result is shown in right. The manipulation relationship between the mobile phone and the box is determined by the following two steps: 1) match detected bounding boxes of the mobile phone and the box to the ground truth ones by overlaps; 2) use manipulation relationship between the ground truth bounding boxes to label the manipulation relationship of detected bounding boxes.

\subsection{Loss Function of Relation Predictor} 

In our work, there are three manipulation relationship types between any two objects in one image:

\begin{itemize}
\item 1) object 1 is the parent of object 2
\item 2) object 2 is the parent of object 1
\item 3) object 1 and object 2 have no manipulation relationship.
\end{itemize}

Therefore, our manipulation relationship prediction process is essentially a classification problem of three categories for any pair of objects $CNN(O_i,O_j;\Phi)$. Let $\Theta$ denote the weights of relation prediction branch. Note that because exchanging the subject and object will possibly change the manipulation relationship type ($e.g.$ from $parent$ to $child$), the prediction of $CNN(O_i,O_j;\Phi)$ and $CNN(O_j,O_i;\Phi)$ may be different. The manipulation relationship likelihood of $R$ is defined as:

\begin{equation}
P(R|O_{i},O_{j};\Theta) = \frac {e^{h_{\Theta}^{R}(CNN(O_i,O_j;\Phi))}} {\sum_{i=1}^{3}e^{h_{\Theta}^{i}(CNN(O_i,O_j;\Phi))}}
\end{equation}

We choose multi-class cross entropy function as loss function of manipulation relationship prediction:

\begin{equation}
L_{rp}(R|O_{i},O_{j};\Theta) =-log(P(R|O_{i},O_{j};\Theta))
\end{equation}

For each image, manipulation relationship prediction loss includes two parts:  online data loss $L_{on}$ and offline data loss $L_{off}$. The loss for the whole image is:

\begin{equation}
\begin{split}
L_{RP}(D_{RP};\Theta) = &\lambda L_{on} + (1-\lambda) L_{off}\\
=&\lambda \sum_{D_{on}} L_{rp}(R|O_{i},O_{j};\Theta) + \\
&(1-\lambda) \sum_{D_{off}} L_{rp}(R|O_{i},O_{j};\Theta) 
\end{split}
\end{equation}

where $\lambda$ is used to balance the importance of online data $D_{on}$ and offline data $D_{off}$. In our work, we set $\lambda$ to 0.5.
 
\subsection{Training Method} 

The whole network is trained end-to-end, which means that the object detector and manipulation relationship predictor are trained simultaneously. 

Let $\Omega$ be the weights of object detector and $D_{OD}$ be the training data of object detector including shared features of the whole image $CNN(I;\Phi)$ and object detection ground truth $(\overline {cls},\overline {loc})$. The loss function for object detector is the same as Liu et al. described in \cite {SSD}:

\begin{equation}
L_{OD}(D_{OD};\Omega) = L_{loc}+\alpha L_{conf}
\end{equation}

where $\alpha$ is set to 1 according to experience. Like in \cite {SSD}, $default$ $bounding$ $boxes$ are defined as a set of predetermined bounding boxes with a few of fixed sizes, which serve as a reference during object detection process. Location loss ${L_{loc}}$ is smooth L1 loss between ground truth bounding box and matched default bounding box and all bounding boxes are encoded as offsets. Classification confidence loss $L_{conf}$ is also multi-class cross entropy loss.

\begin{algorithm}[t]
\caption{Training Algorithm} 
\label {alg1}
\hspace*{0.02in} {\bf Input:} 
Training set of images with object location $\overline {loc}$, category $\overline {cls}$  and manipulation relationship $(O_i,R,O_j)$, pretrained VGG \cite {c6} or ResNet50 parameters \cite {c7}\\
\hspace*{0.02in} {\bf Output:} 
Feature extractor $\Phi$, object detector $\Omega$ and manipulation relationship predictor $\Theta$
\begin{algorithmic}[1]
\State Initialize feature extractor $\Phi$ using pretrained VGG or ResNet50 and object detector $\Omega$, manipulation relationship predictor $\Theta$ randomly.
\State Pretrain feature extractor $\Phi$ and object detector $\Omega$ for 10k iterations on images\cite {SSD}.
\While{$iter<maxiter$}
  \State Randomly sample a mini-batch.
  \State Extract CNN features $CNN(I;\Phi)$
  \State Detect objects and get a set of online predicted bounding boxes $B_{on}$
  \State Get offline bounding boxes $B_{off}$
  \State Feed $\{CNN(I;\Phi),B_{on}\}$ and $\{CNN(I;\Phi),B_{off}\}$ into OP$^2$L and get manipulation relationship training data $D_{RP}$
  \State Update object detector $\Omega$ and manipulation relationship predictor $\Theta$ using $\{D_{OD},L_{OD}\}$ and $\{D_{{RP}},L_{RP}\}$
  \State Update feature extractor using $L(I;\Phi)$
\EndWhile
\State \Return $\Phi,\Omega,\Theta$
\end{algorithmic}
\end{algorithm}

Loss function of manipulation relationship prediction $L_{RP}$ is detailed in section IV.B. Combining $L_{RP}$ and $L_{OD}$, the complete loss for shared layers is:
 
\begin{equation}
\begin{split}
L(I;\Phi) = \mu L_{OD}(D_{OD};\Omega) + 
&(1-\mu) L_{RP}(D_{RP};\Theta)
\end{split}
\end{equation}
 
$\mu$ is used to balance the importance of $L_{OD}$ and $L_{RP}$. In our work, $\mu$ is set to 0.5. And according to chain rule:
\begin{equation}
\begin{split}
\frac {\partial L}{\partial \Phi} =& \mu \frac {\partial L_{OD}} {\partial CNN(I;\Phi)} \frac {\partial CNN(I;\Phi)}{\partial \Phi} +\\
&(1-\mu) \frac {\partial L_{RP}}{\partial CNN(O_i,O_j;\Phi)} \frac {\partial CNN(O_i,O_j;\Phi)}{\partial \Phi} 
\end{split}
\end{equation}

Complete training algorithm is shown in Algorithm \ref{alg1}.

\section{Dataset}

\subsection {Data Collection}

Different from visual relationship dataset\cite {c1}, we focus on manipulation relationships, so objects included in our dataset should be manipulatable or graspable. Moreover, manipulation relationship dataset should contain not only objects localized in images, but also rich variety of position relationships.

\begin{figure}[tb]
\centering
\subfigure[Category and manipulation relationship distrubution]{
\label{fig:subfig:a} 
\includegraphics[scale=0.3]{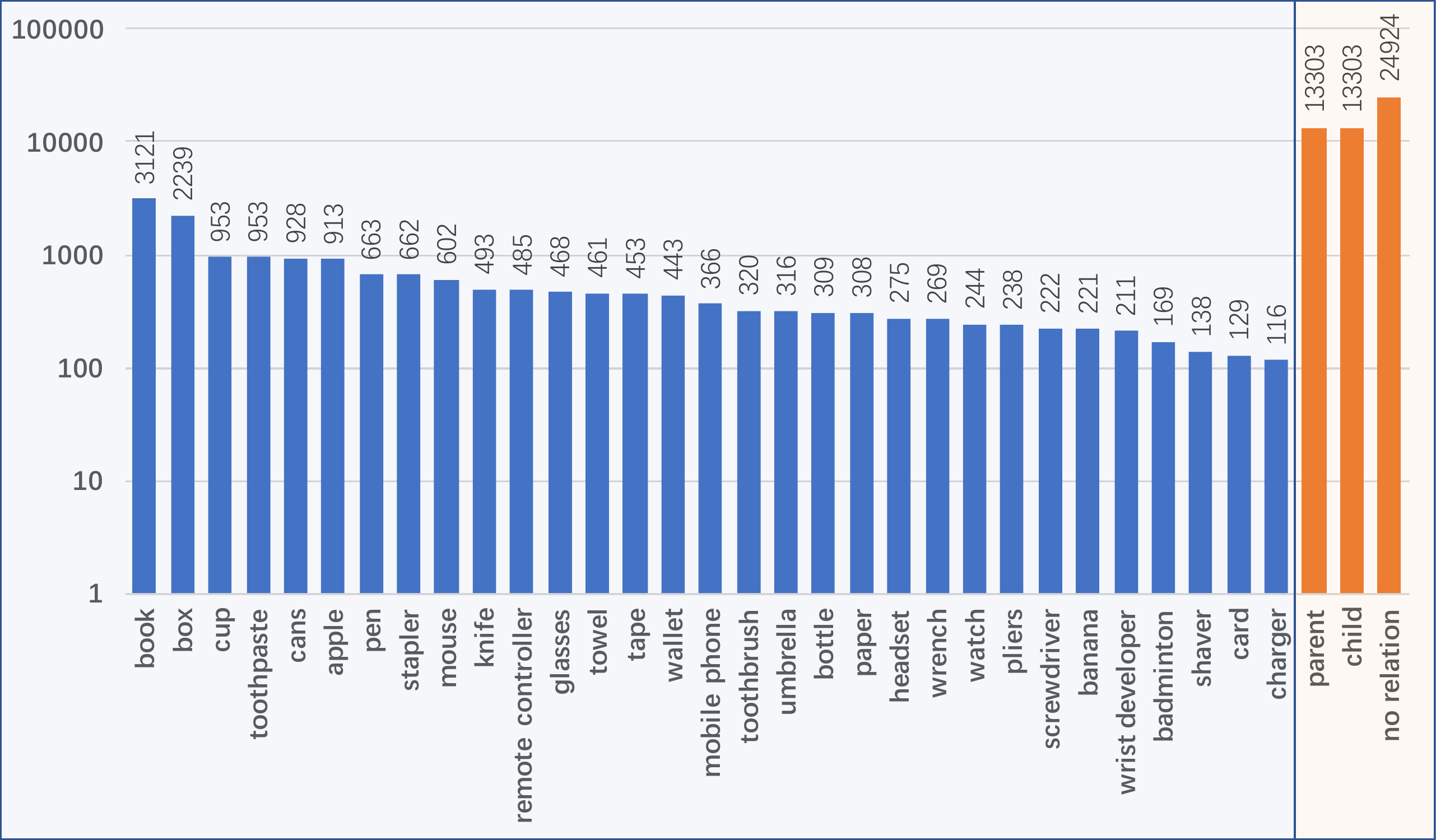}}
\subfigure[Dataset examples]{
\label{fig:subfig:b} 
\includegraphics[scale=0.1]{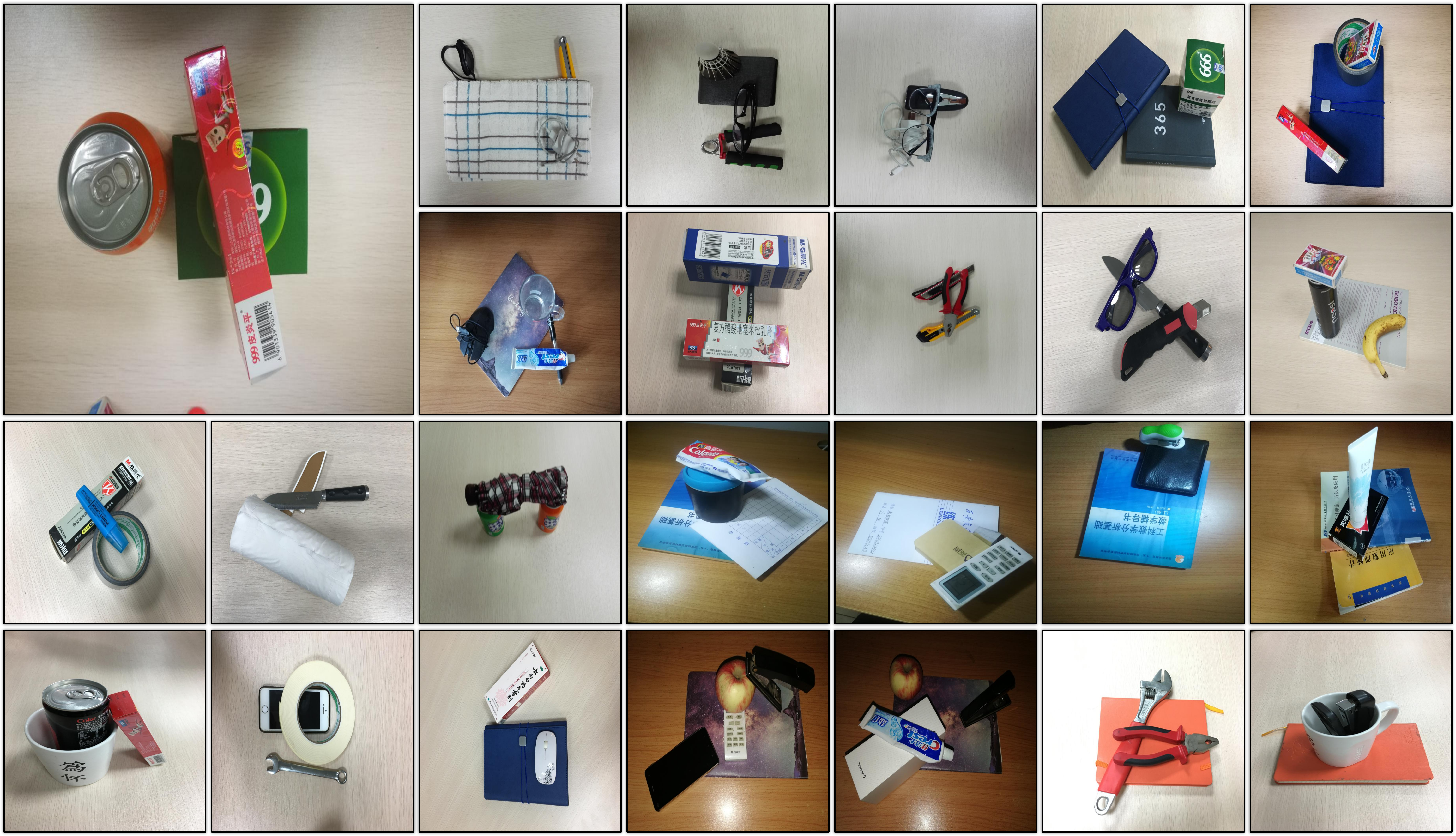}}
\caption{Visual Manipulation Relationship Dataset. (a) Category and manipulation relationship distribution of our dataset. (b) Some dataset examples}
\label{fig:subfig} 
\end{figure}

Our data are collected and labeled using hundreds of objects coming from 31 categories. There are totally 5185 images including 17688 object instances and 51530 manipulation relationships. Category and manipulation relationship distribution is shown in Fig. \ref{fig:subfig:a}. The annotation format is similar to PASCAL VOC: each object node includes category information, bounding box location, different from PASCAL VOC, the index of the current node and indexes of its parent nodes and child nodes. Some examples of our dataset is shown in Fig. \ref{fig:subfig:b}.

During training, we randomly split the dataset into a training set and a testing set in a ratio of nine to one. In detail, training set includes 4656 images, 15911 object instances and 46934 manipulation relationships, and testing set contains the rest.

\begin{table*}[h]
\caption{Accuracy of Object Detection and Visual Manipulation Relationship Prediction}
\label{resultstable}
\begin{center}
\begin{tabular}{|l|l|l|c|c|cc|c|c|}
\hline
\bf {Author} & \bf {Algorithm} & \bf{Training Data} & \begin{minipage}{1cm} \vspace{0.1cm}\centering \bf {mAP} \vspace{0.1cm}\end{minipage} & \bf{Rel.} & \bf{Obj.Rec.} & \bf{Obj.Prec.} & \bf{Img.} & \begin{minipage}{1.5cm} \centering \bf{Speed (ms)} \end{minipage}  \\
\hline
\hline
\bf{Lu et al.}\cite{c1} &\begin{minipage}{3.7cm} \vspace{0.1cm} VGG-SSD, VAM \vspace{0.1cm}\end{minipage}  &$D_{on}\cup D_{off}$&93.01 & 88.76  & 75.50 & 71.28 & 46.88 & \multirow{2}{*}{$>$100}\\\
&\begin{minipage}{3.7cm} \vspace{0.1cm} ResNet-SSD, VAM\vspace{0.1cm}\end{minipage}  &$D_{on}\cup D_{off}$                              &91.72 & 88.76 & 74.33 &75.19 & 49.72 & \\
\hline
\hline
\bf{Ours}&\begin{minipage}{3.7cm} \vspace{0.1cm} VGG-VMRN (No Rel. Grad.) \vspace{0.1cm}\end{minipage}&$D_{on}\cup D_{off}$& 93.01       & 88.36        & 77.28      & 73.04      & 50.66        & \\
&\begin{minipage}{3.7cm} \vspace{0.1cm} ResNet-VMRN (No Rel. Grad.)\vspace{0.1cm}\end{minipage}           &$D_{on}\cup D_{off}$& 91.72       & 90.73        & 77.68       & 77.55      & 53.12        &\\
&\begin{minipage}{3.7cm} \vspace{0.1cm} VGG-VMRN\vspace{0.1cm}\end{minipage}                                        & $D_{on}$                  & 94.18       & 92.80  & \bf{82.64}      & 77.76       & 60.49        & \\
& \begin{minipage}{3.7cm} \vspace{0.1cm} VGG-VMRN\vspace{0.1cm}\end{minipage}                                       &$D_{off}$                   & \bf{94.36} &  92.75      &   81.55     & 76.09       & 58.60       &  \multirow{2}{*}{\bf{28}} \\
&\begin{minipage}{3.7cm} \vspace{0.1cm} VGG-VMRN\vspace{0.1cm}\end{minipage}                                        &$D_{on}\cup D_{off}$ & 94.09      &\bf{93.36}  & 82.29            & \bf{78.01}       &\bf{63.14}   & \\
&\begin{minipage}{3.7cm} \vspace{0.1cm} ResNet-VMRN\vspace{0.1cm}\end{minipage}                                    &$D_{on}$                   & 91.81        & 92.01       & 79.03           & 72.55       & 54.44     &  \\
&\begin{minipage}{3.7cm} \vspace{0.1cm} ResNet-VMRN\vspace{0.1cm}\end{minipage}                                    &$D_{off}$                    & 92.71        & 91.86       & 79.33           & 74.71     & 55.95     &  \\
&\begin{minipage}{3.7cm} \vspace{0.1cm} ResNet-VMRN\vspace{0.1cm}\end{minipage}                                    &$D_{on}\cup D_{off}$ & 92.67        & 92.19       & 80.55           & 76.02     &  57.28     &  \\
\hline
\end{tabular}
\end{center}
\end{table*}

\subsection {Labeling Criterion}

Because our dataset focuses on the manipulation relationship with no linguistic or position information, instead of directly giving position relationships ($e.g.$ under, on, beside and so on) between objects, we only give the order of manipulation of objects: manipulation relationship tree. There are several advantages over giving position relationships: 1) the manipulation relationships are more simpler, which makes relationship prediction task easier; 2) the prediction can directly give the manipulation relationships between objects, without the need to reconstruct the manipulation relationships through position relationships. 

During labeling, there should be a criterion that can be strictly enforced. Therefore, in our work, we set a {\bf labeling criterion} of manipulation relationship: when the movement of an object will affect the stability of other objects, the object should not be the leaf node of the manipulation relationship tree, which means that the object should not be moved first. For example, as shown in the up-left image in Fig. \ref{fig:subfig:b}, there are three objects: on the left, there is an orange can and on the right, a red box is put on a green box. If the green box is moved first, it will have an effect on the stability of the red box, so it should not be the leaf node of the manipulation relationship tree. If the red box or the can is moved first, it will not affect stability of any other object, so they should be the leaf node.

\section{Experiments}

\subsection{Training Settings}
 
Our models are trained on Titan Xp with 12 GB memory. We have trained two Visual Manipulation Relationship Network (VMRN) models based on VGGNet and ResNet called VGG-VMRN and ResNet VMRN. Because of the unstability of the random object detection results in the beginning, the two VMRN models are pretrained without $D_{on}$ for the first 10k iterations. Detailed training settings are listed in Table \ref{trainsettings}.
 
 \begin{table}[t]
\caption{Training Settings}
\label{trainsettings}
\begin{center}
\begin{tabular}{|c|c|c|}
\hline
\bf{Hyperparam.}  & \begin{minipage}{2cm}\vspace{0.1cm} \centering \bf{VGG-VMRN} \vspace{0.1cm}\end{minipage} & \begin{minipage}{2cm} \centering \bf{ResNet-VMRN}\end{minipage}\\
\hline
\hline
\multirow{2}{*}{Learning Rate} & 0-80k iters: 1e-3 & 0-80k iters: 1e-3 \\
& 80k-120k iters: 1e-4 & 80k-120k iters: 1e-4 \\
\hline
Learning Rate Decay& 0 & 0\\
\hline
Weight Decay & 3e-3 & 1e-4 \\
\hline
Batch Size & 8 & 8 \\
\hline
Momentum & 0.9 & 0.9 \\
\hline
Nesterov & True & True \\
\hline
Max Epoches & 120 & 120 \\
\hline
Iters per Epoch & 1000 & 1000 \\
\hline
Framework & Torch7 & Torch7 \\
\hline
\end{tabular}
\end{center}
\end{table}

 \begin{figure*}[t]        
 \center{\includegraphics[scale=0.28]{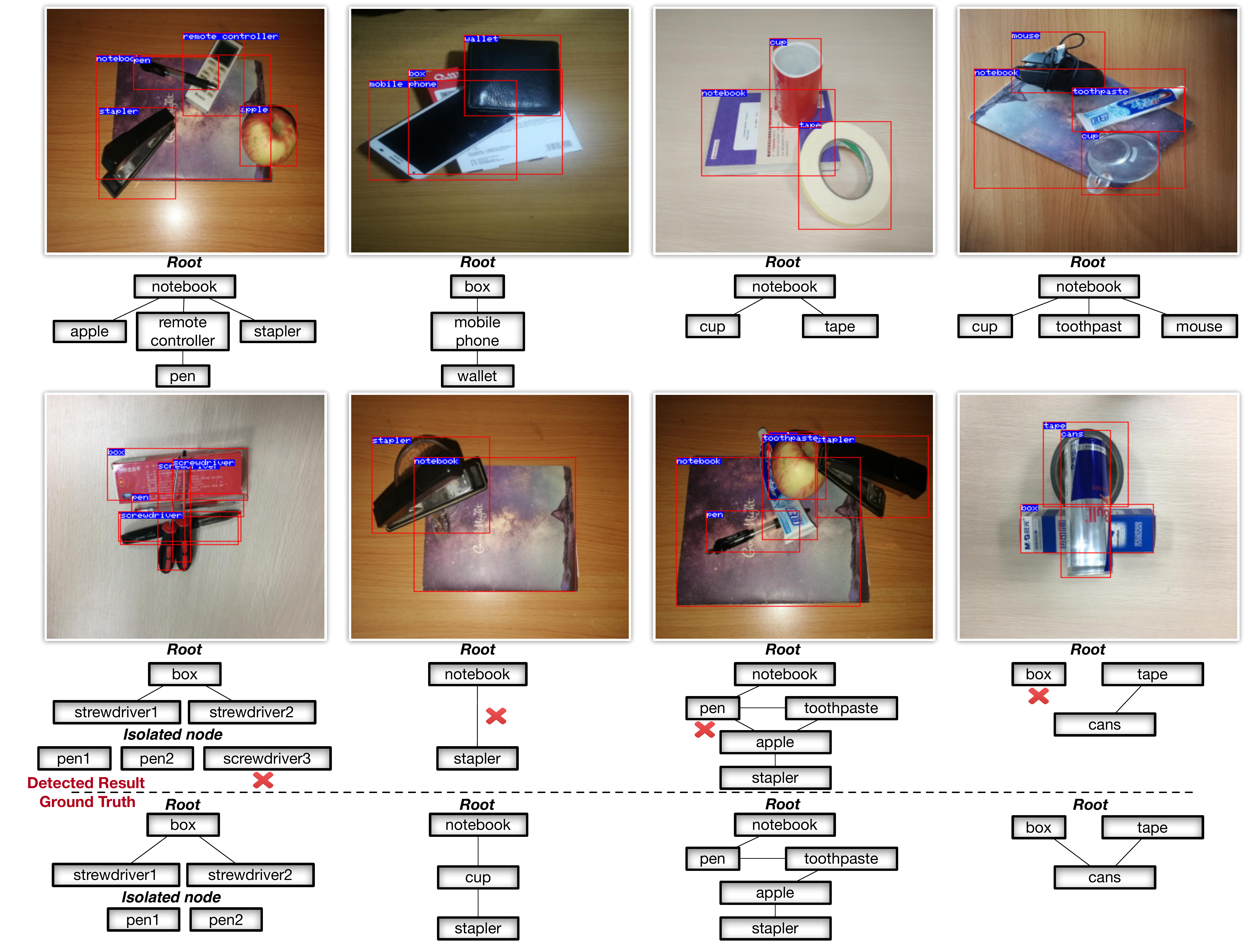}}        
 \caption{\label{fig6} Result examples. Upper: examples with right object detection and manipulation relationship prediction. Lower: examples with wrong results (from left to right: redundant object detection, failing object detection, redundant manipulation relationship, failing manipulation relationship)}      
 \end{figure*}

\subsection{Testing Settings}
 
{\bf Comparison Model} As we know, there is no research about robotic manipulation relationships so far. So we compare our experiment results with Visual Appearance Model (VAM) in Lu et al. \cite{c1}, which is modified to adapt to our task. VAM takes union bounding box as input and outputs the relationship. But in our work, exchanging the subject and object may change the manipulation relationship. Therefore, instead of only using union bounding box, we parallel subject, object and union bounding boxes as input to get the final manipulation relationship.

{\bf Self Comparison} To study the contribution of OP$^2$L and end-to-end training, we also confirm the performance of our models that are trained with no gradients backward from manipulation relationship predictor (\{VGG-SSD, VMRN (No Rel. Grad.)\} and \{ResNet-SSD, VMRN (No Rel. Grad.)\}). To explore the benefits from online and offline data, we also train our models with only online ($D_{on}$) or offline ($D_{off}$) training data.
 
{\bf Metrics} Three metrics are used in our experiment: 
1) Manipulation Relationship Testing (Rel.): this metric focus on the accuracy of manipulation relationship model on ground truth object instance pairs, in which the input features or image patches of manipulation relationship predictor are obtained based on the offline ground truth bounding boxes; 2) Object-based Testing (Obj. Rec. and Obj. Prec.): this metric tests the accuracy based on object pairs. In this setting, the triplet $(O_i,R,O_j)$ is treated as a whole. The prediction is considered correct if both objects are detected correctly (category is right and IoU between predicted bounding box and ground truth is more than 0.5) and the predicted manipulation relationship is correct. We compute the recall (Obj. Rec.) and precision (Obj. Prec.) of our models during object-based testing 3) Image-based Testing (Img.): this metric tests the accuracy based on the whole image. In this setting, the image is considered correct only when all possible triplets are predicted correctly.

\subsection{Analysis}

Results are shown in Table \ref{resultstable}. Compared with VAM, we can conclude that: 

1) {\bf Performance is better:} VAM performs worse than proposed VMRN models in all three experiment settings. The gains mainly come from the end-to-end training process, which improves the accuracy of manipulation relationship prediction a lot (from 88.76\% to 93.36\%). This is confirmed in the following self comparison part.

2) {\bf Speed is faster:} The proposed VMRN models (VGG-VMRN and ResNet-VMRN) are both less time-consuming than VAM. Forward process of OP$^2$L and manipulation relationship predictor takes 5.5ms per image in average. As described in \cite{SSD}, the speed of SSD object detector is 21.74ms per image on Titan X with mini-batch of 1 image. So our manipulation relationship prediction has little effect on speed of the whole network. But because of the huge network architecture and sequential prediction process, VAM spends 122ms on each image in average to predict all of the manipulation relationships. Even when we put all possible triplets of one image to a batch, it still spends 86ms for one image.

Self comparison results indicate that proposed VMRN models trained end-to-end can outperform the models that trained without the gradients from manipulation relationship predictor. It mainly benefits from the influence coming from manipulation relationship prediction loss $L_{RP}$. The parameters of the network are adjusted to better predict the visual manipulation relationships and the network is more holistic. As explored in Pinto et al.\cite{multilearn}, multi-task learning in our network can help improve the performance because of diversity of data and regularization in learning. Finally, we can observe that using online and offline data simultaneously may actually help to improve the performance of the network due to the complementing of online and offline data.

The difference between the performance of VGG-VMRN and ResNet-VMRN is also interesting. Gradients coming from manipulation relationship prediction loss $L_{RP}$ improve both networks, but its improvement on ResNet-VMRN is less than that on VGG-VMRN as shown in Table \ref{resultstable}. Note that VGG-based feature extractor has 7.63 million parameters and ResNet-based feature extractor has 1.45 million parameters, so the number of parameters may limit the performance ceiling of ResNet-VMRN. In the future, we will try deeper ResNet as our base network.

Some subjective results are shown in Fig. \ref{fig6}. From the four examples in the first line, we can see that our model can simultaneously detect objects and manipulation relationships in one image. From the four examples in the second line, we can conclude that the occlusion, the similarity between different categories and visual illusion can have a negative influence on the predicted results.

 
\section{Conclusions}

In this paper, we focus on solving the problem of visual manipulation relationship prediction to help robot manipulate things in the right order. We propose a new network architecture named Visual Manipulation Relationship Network and collect a dataset called Visual Manipulation Relationship Dataset to implement simultaneously detecting objects and predicting manipulation relationships, which meets the real-time requirement on robot platform. The proposed Object Paring Pooling Layer (OP$^2$L) can not only accelerate the manipulation relationship prediction by replacing the sequential prediction with a simple forward process, but also improve the performance of the whole network by back-propagating the gradients from manipulation relationship predictor.

However, due to the limited number of objects used in training, it is difficult for the object detector to generalize to objects with a large difference in appearance from our dataset. In our future work, we will expand our dataset using more graspable objects and combine the grasp detection with VMRN to implement an all-in-one network which can simultaneously detects objects and their grasp positions and predicts the correct manipulation relationships.

\addtolength{\textheight}{0cm}   



\section*{ACKNOWLEDGMENT}

This work was supported in part by NSFC No. 91748208, the National Key Research and Development Program of China under grant No. 2017YFB1302200 and 2016YFB1000903,  NSFC No. 61573268, and Shaanxi Key Laboratory of Intelligent Robots.

\bibliographystyle{unsrt}
\bibliography{IROS2018}

\end{document}